\documentclass[11pt]{article}

\usepackage[preprint]{acl}
\usepackage{times}
\usepackage{latexsym}
\usepackage{booktabs}
\usepackage[T1]{fontenc}
\usepackage{pifont}
\usepackage{amssymb}
\usepackage{enumitem}
\usepackage{microtype}
\usepackage{inconsolata}
\usepackage{graphicx}
\usepackage{array}
\usepackage{tabularx}
\newcolumntype{L}{>{\raggedright\arraybackslash}X}
\usepackage[table]{xcolor}
\usepackage{colortbl}
\usepackage{heatcell}
\usepackage[most,breakable]{tcolorbox}
\usepackage{fvextra}

\newtcolorbox{promptbox}[1]{
  enhanced, breakable, colback=gray!4, colframe=black!55,
  colbacktitle=black!75, coltitle=white,
  fonttitle=\bfseries\small\sffamily,
  title={#1},
  boxrule=0.4pt, arc=1.5pt, left=4pt, right=4pt, top=3pt, bottom=3pt,
  before skip=6pt, after skip=6pt,
}
\fvset{fontsize=\footnotesize, breaklines=true, breakanywhere=true,
       breaksymbolleft=\hbox{}, breaksymbolright=\hbox{}}

\newcommand{\CN}{Real }

\title{Closing Ambient Clinical Documentation Gaps\\with Automated Provider Queries}
\author{
Joseph Paul Cohen \\ Amazon  \\ \texttt{\small joseph@josephpcohen.com} \\\And
Raj Shah \\ Amazon   \\\And
Han-Chin Shing  \\ Amazon  \\\And
Fang Wang  \\ Amazon  \\\AND
Susan Nguyen, RN \\ Amazon \\\And
Chaitanya Shivade \\ Amazon \\\And
Jack Moriarty \\ Amazon  
}

\begin{document}
\maketitle

%% ============================================================
\begin{abstract}

Provider queries are clarifying requests sent by clinical documentation specialists to physicians to close gaps in the clinical note and ensure accurate billing. Prior work automates note drafting, ICD-10 coding, and order extraction assuming a complete transcript, leaving these gaps unaddressed. We study whether an LLM can automate the query loop, termed DAU (Draft, Ask, Update), across those three tasks. An audit of 3,000 real visits identifies the sources of missing documentation, from which we build five transcript-degradation benchmarks on public data. Analyzing 21k clarification turns on real conversations, we find useful-question predictors are task-specific: oracle confidence dominates, but note completeness needs only simple recall questions while ICD-10 coding needs harder, multi-option ones. About 9\% of turns hurt performance, driven by redundant questions and non-answers that still trigger a rewrite. Deployment depends on learning ``when not'' as much as ``what to'' ask.

\end{abstract}

%% ============================================================
\section{Introduction}

Clinical encounter transcripts are routinely incomplete: implicit reasoning, paraphrased jargon, and transcription errors leave gaps that propagate into the note, the ICD-10 codes, and the order list \cite{carandang2025llms}. An incomplete note can lead to miscoded diagnoses, mis-billed visits, and errors in quality and safety reporting. Hospitals have clinical documentation integrity (CDI) programs to catch issues and route queries back to physicians \cite{reyes2017cdi}. Answering those queries takes physician time and is a known driver of documentation burden and burnout \cite{rouse2022cdi}. Automating part of this loop helps both record accuracy and physician time.

\begin{figure*}[t]
\centering
\includegraphics[width=\linewidth]{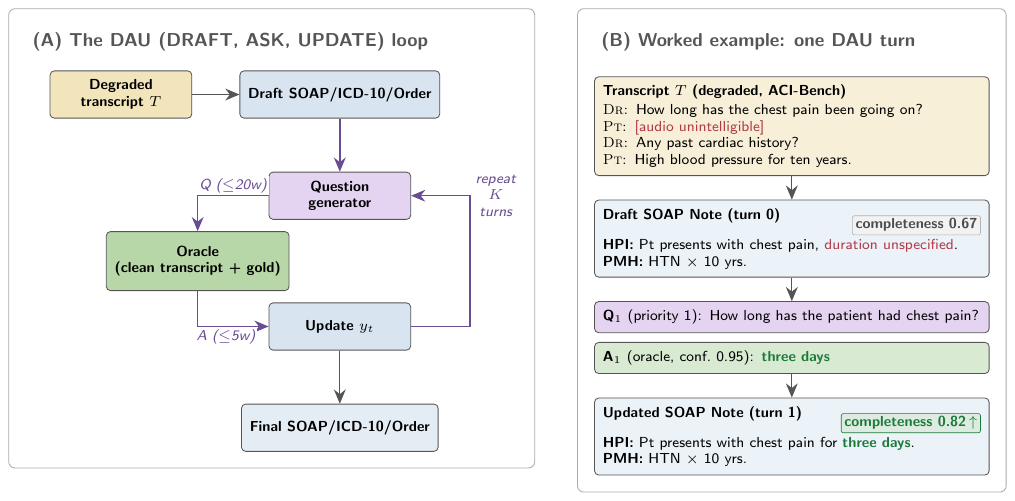}
\caption{Overview of DAU (Draft, Ask, Update).
\textbf{(A)} A single iterative loop (Draft, Ask, Update) is shared across three clinical tasks (note generation, ICD-10 coding, order extraction). The system sees only the (possibly degraded) transcript $T$; the simulated-clinician oracle sees the original transcript and the gold reference. Questions are capped at 20 words and answers at 5.
\textbf{(B)} A single DAU turn on a degraded ACI-Bench dialogue: the patient turn stating chest-pain duration is unintelligible, so the draft note has a gap. One clarifying question and a 5-word oracle answer recover the missing fact and raise note completeness from 67 to 82.}
\label{fig:overview}
\end{figure*}

We study how to automate provider queries using a \emph{DAU} loop (Draft, Ask, Update) (Figure \ref{fig:overview}) instantiated for three downstream tasks (note generation, ICD-10 coding, and medical order extraction) and study what makes a good question.

\begin{table}[t]
\centering
\small
\setlength{\tabcolsep}{4pt}
\definecolor{softrose}{RGB}{200,150,175}
\newheatcol{g}{min=0, max=33, color=softrose}
\begin{tabularx}{\columnwidth}{@{}lgL@{}}
\toprule
\textbf{Cause} & \textbf{\%} & \textbf{Description} \\
\midrule
Chart Context    & 32.9 & Lives only in the EHR (doses, history, demographics) \\
Visual/Physical  & 32.0 & Exam findings and vitals never spoken aloud \\
Drafter Omission & 14.9 & In the transcript but dropped by the drafter \\
Implied Context  & 11.2 & Defaulted negatives; vague back-references \\
Other            &  5.6 & Residual \\
Patient-facing   &  2.6 & Lay phrasing (``labs'') where the EHR is specific \\
ASR              &  0.8 & Number, dose, or entity garbled \\
\bottomrule
\end{tabularx}
\caption{Causes of missing facts in baseline LLM-drafted notes on $N{=}3{,}000$ real ambient clinical conversations (LLM-assigned categories). Nearly half of the gap (Visual/Physical, Implied Context, Patient-facing, ASR) is answerable by a short clinician query.
}
\label{tab:cohort-gap-intro}
\end{table} 

\paragraph{What does the gap look like in practice?}
To motivate this work on real clinical data, we audited the missing facts in baseline LLM-drafted notes on a private, internal sample of 3{,}000 primary-care visits from US hospitals, ASR-transcribed from ambient audio with EHR-derived gold notes. All records were de-identified and anonymized in accordance with HIPAA standards.
Draft notes were produced from audio transcripts with the \citet{abacha2023empirical} SOAP prompt. Each fact missing from the draft was passed to an LLM auditor that read the ASR transcript and assigned a cause from the rubric in Table~\ref{tab:cohort-gap-intro}. These categories were determined by an LLM and curated by clinical experts.

The reason for these errors split into three categories: 1. requiring information from the clinician, 2. from the EHR, or 3. improved summarization process. The clinician is required for \emph{Visual/Physical} (a one-word answer reports the exam finding), \emph{Implied Context} (the unspoken default becomes explicit), \emph{Patient-facing} (the lay term is replaced with the EHR-precise one), and \emph{ASR} (the garbled token is re-stated) errors, together $\sim$$47\%$ of the gap. 
The other major slices, \emph{Chart Context} ($33\%$) and \emph{Drafter Omission} ($15\%$), are $\sim$$48\%$ combined, are likely addressed by improving the automatic summarization process.

%\newpage
\paragraph{Contributions:}
\begin{itemize}[noitemsep,topsep=2pt,leftmargin=*]
    \item Quantify sources of note completeness gaps on $3{,}000$ real encounters, isolating a $\sim$47\% slice a clarification loop can plausibly recover.
    \item Demonstrate a DAU (Draft, Ask, Update) loop across three clinical tasks: note completeness, ICD-10 coding, and medical order extraction.
    \item Evaluate on public datasets (ACI-Bench, PriMock57, MIMIC-IV, MEDIQA-OE).
    \item Create and share a proxy dataset with five targeted transcript degradations (ASR noise, dropped content, implied context, patient-facing paraphrase, visual/physical signals).
    \item Empirical analysis of question usefulness on real clinical conversations, we identify oracle answer confidence and question redundancy as the dominant predictors of per-turn metric change.
\end{itemize}

%% ============================================================
\section{Related Work}

\paragraph{Provider queries and clinical documentation improvement.}
The healthcare workflow that DAU automates is the \emph{provider query}: when a clinical documentation integrity (CDI) specialist or coder finds a gap in a note, they send the provider a written question to clarify the diagnosis, severity, or other coding-relevant detail \cite{ahima2013query}. These queries themselves are a known source of physician burden and workflow friction \cite{rouse2022cdi}. The closest non-LLM analogues to DAU are real-time EHR-rule tools such as AutoDx \cite{cerasale2024autodx,smith2025autodx}, which fire pre-authored prompts to providers and reduce CDI query volume by 50--60\%; Our DAU approach differs in that it generates the question from the transcript and current draft rather than from a hand-coded rule set, and operates over notes, codes, and orders within one loop.

\paragraph{Clarifying questions.}
Asking clarifying questions has been studied in conversational search \cite{aliannejadi2019asking}, community QA \cite{rao2018learning}, and interactive clinical reasoning \cite{li2024mediq,li2025alfa}.
Closely related, \citet{hu2024uot} use uncertainty-based rewards to drive LLM information-seeking in medical diagnosis, and \citet{tu2025amie} optimizes a single LLM for diagnostic history-taking via simulated self-play.

\paragraph{Clinical note generation.}
Early neural approaches generated notes from dialogue with sequence-to-sequence models \cite{enarvi2020generating}.
ACI-Bench \cite{yim2023acibench} and PriMock57 \cite{korfiatis2022primock57} are the two widely-used public doctor--patient dialogue datasets; MEDIQA-Chat \cite{abacha2023empirical} established a SOAP prompt still used as the LLM zero-shot baseline.
\citet{vanveen2024adapted} benchmark adapted LLMs across radiology, progress notes, and doctor--patient dialogue, finding LLM summaries match or exceed expert summaries on completeness and correctness.
Recent work tunes open-source models for the same task \cite{wang2024towards} and documents reliability issues in LLM-generated notes \cite{carandang2025llms}.
All of these systems assume the transcript is complete.

\paragraph{Automated ICD coding.}
Supervised ICD coders are mostly trained on MIMIC-III/IV including CAML \cite{mullenbach2018caml}, LAAT \cite{vu2020laat}, MultiResCNN \cite{li2020multirescnn}, and PLM-ICD \cite{huang2022plmicd}; Edin et al. \cite{edin2023automated} reproduce them under a uniform MIMIC-IV ICD-10 split.
LLM Prompt based methods make it easier to utilize larger models however single-shot LLM coders have been found to hallucinate invalid codes \cite{soroush2024llmcoders} motivating decompose-then-map pipelines \cite{boyle2023automatedcoding} and retrieval-augmented candidates \cite{yang2022kept}.
None of this work addresses underspecified dialogue inputs; workflow reviews \cite{gan2024aligning,bi2024harmonising} argue this gap matters clinically.

\paragraph{Medical order extraction.}
The MEDIQA-OE 2025 shared task \cite{corbeil2025mediqaoe} targets extraction of medications, labs, imaging, and follow-ups from encounter dialogues and supplies the zero-shot prompt we use as a baseline.

\paragraph{Controlled underspecification.}
\citet{bn2025factcontrol} introduce Leave-$N$-Out for fact-controlled hallucination evaluation.
We reuse that philosophy but target dialogue-level degradations (\S\ref{sec:degradations}) so that the resulting notes, codes, and orders can all be scored against the same gold references.

\paragraph{Our contribution.} Prior clarification systems either use hand-authored EHR rules (AutoDx) or drive patient-facing diagnostic dialogue (MediQ, AMIE, UoT); we instead generate provider-facing questions grounded in the recorded transcript and current draft. Prior note-generation, ICD-coding, and order-extraction systems all assume the transcript is complete; we treat it as underspecified and recover missing facts at inference time without retraining, in a single loop that spans all three tasks. Unlike Leave-$N$-Out \cite{bn2025factcontrol}, which perturbs isolated facts, our five ACI-Bench degradations act at the dialogue level so the same encounter is scored jointly on notes, codes, and orders.

%% ============================================================
\section{Method}

\begin{table*}[t]
\centering
\small
\setlength{\tabcolsep}{6pt}
\begin{tabular}{@{}l p{5.2cm} p{3.5cm} p{3.5cm}@{}}
\toprule
\textbf{Degradation Type} & \textbf{What it simulates} & \textbf{Original} & \textbf{Degraded} \\
\midrule
ASR              & Phonetic, sound-alike transcription errors in clinical terms.        & \textit{lipid}; \textit{Losartan}             & \textit{livid}; \textit{Lorazepam} \\
Dropped Content  & Clinically-meaningful turns removed from the transcript entirely.    & \textit{``I've been having chest pain for a week.''} & (turn removed) \\
Implied Context  & Explicit facts replaced with vague back-references to shared history.& \textit{Metformin 500\,mg BID}                & \textit{same as last time} \\
Patient-facing   & Doctor turns reworded in lay language (drugs, labs, procedures).     & \textit{order a lipid panel}                  & \textit{run some bloodwork} \\
Visual/Physical  & Spatial / physical details replaced with deictic references.        & \textit{8\,cm; left hand}                     & \textit{about this much; this side} \\
\bottomrule
\end{tabular}
\caption{The five transcript degradation types applied at 20\% to every dialogue dataset. Non-drop types enforce a same-length-or-shorter substitution constraint; dropped turns disappear from the final transcript. Each sample retains an audit trail of changes for provenance.}
\label{tab:degradations}
\end{table*}

\subsection{Transcript Degradations}
\label{sec:degradations}

Real clinical transcripts fail in distinct, characterizable ways: speech-recognition systems mishear drug names, patients leave information implicit, and visual cues never make it into the audio. To evaluate downstream tasks under these failure modes in a controlled way, we apply five 20\% perturbation variants to transcripts from the ACI-Bench and PriMock57 datasets. These perturbation types were determined from our analysis of real errors as well as interviewing a clinical scribe. Table \ref{tab:degradations} summarizes each type with a before/after example. This transformation was done by prompting DeepSeek V3.2 in order to comply with corporate policies and allow us to release the data.

\subsection{The DAU Loop}
\label{sec:dau-loop}

In order to identify realistic questions and to craft the question generation prompts we interviewed a clinical scribe and enumerated real clinical questions (see \S\ref{appendix:real-queries})

Given a transcript $T$ that is used to generate some output $y$ (note, code set, or order list) and a budget $k$ of how many questions can be asked:

\paragraph{Turn 0.} Generate $y_0$ from $T$ with a task-specific zero-shot prompt.

\paragraph{Turns $1\dots k$.} Repeat up to the budget $k$:
\begin{enumerate}[noitemsep,topsep=2pt]
    \item \textbf{Question generation.} Given $T$, $y_{t-1}$, and previous Q\&A, produce one clarifying question targeting the most impactful missing or uncertain element, or abstain and end the loop. Questions are capped at 20 words in length
    \item \textbf{Oracle answer.} A simulated clinician with access to the original transcript/summary as well as the ground truth facts; answers are capped at 5 words.
    \item \textbf{Output update.} Regenerate $y_t$ by applying only the new information from the Q\&A, preserving unaffected content.
\end{enumerate}

%% ============================================================
\section{Experiments}

We report results per task on the test split of each dataset, running the full DAU loop at budgets $\{0, 1, 3, 5\}$. Budget=$0$ means zero clarification questions are asked, differing from the zero-shot baselines only in the prompt wording, not in the information available. 
DAU Budget=$\infty$ is an upper bound on what any clarification method can achieve under our pipeline. We feed the gold reference 
into the same drafter / coder / extractor and the same evaluator. The gap from 100 measures model generation and evaluator noise rather than missing information.
Detailed dataset information is included in Appendix \ref{sec:appendix-datasets}. 

Baselines are included to illustrate that our Budget=$0$ is not a straw-man that is easily improved. All baselines are run using the Claude Opus 4.6 backbone to isolate the prompt from the backbone.

\subsection{Clinical Note Completeness}
\label{sec:results-notes}

The model transforms a transcript into a SOAP note using the prompt of \citet{abacha2023empirical}. The notes are then scored by fact recall where an LLM auditor extracts a list of atomic individually-checkable facts from the gold note and a second LLM call checks each fact against the generated note \cite{moramarco2022human}. 
We evaluate two zero-shot SOAP prompts as baselines \citep{yim2023acibench,abacha2023empirical}.

Table \ref{tab:notes} reports fact-level completeness. Completeness increases monotonically with budget. The DAU Budget=$\infty$ upper bound sits at $\geq$97.4 on every column.

\subsection{ICD-10 Coding}
\label{sec:results-coding}

The model emits a set of ICD-10 codes scored against the gold codes by micro-F1 following the literature. The DAU oracle is given the gold codes and their canonical long descriptions, with any ICD-10-shaped word redacted from its replies so it cannot leak codes directly.

As baselines we include two LLM coders (two-step \citep{boyle2023automatedcoding}, a Retrieval-coder \citep{yang2022kept}), and three supervised MIMIC-IV models (CAML, LAAT, MultiResCNN) loaded from the \citet{edin2023automated} checkpoints.

Table \ref{tab:coding} reports F1.
On ACI-Bench notes DAU improved Implied Context, exactly the degradations where the draft note is most likely to underspecify; the supervised MIMIC coders have low F1 here likely because the content distribution (single-encounter outpatient dialogue) differs from MIMIC-IV discharge summaries. 
On ICD-Synth, where the input already spells out diagnoses verbatim, Retrieval-coder wins because the coder only has to match explicit phrases.
DAU produces consistent results across all four input distributions. 
DAU Budget=$\infty$ reaches only 44 on ICD-Synth-All. This is because the model was only given the descriptions of the codes and not the codes themselves. The gold codes are still emitted as strings by the LLM, which on the full $\sim$70k-code ICD-10 vocabulary loses substantial F1 to description$\rightarrow$code-string mapping noise (it hallucinates close-but-wrong codes like \texttt{M54.50} vs. gold \texttt{M54.5}); Retrieval-coder bypasses this step (looking codes up in a description index) and is therefore higher than Budget=$\infty$.

\subsection{Medical Order Extraction}
\label{sec:results-orders}

The model emits a list of medical orders (medications, labs, imaging, follow-ups); outputs are scored against the MEDIQA-OE 2025 gold orders by Description Match-F1 \cite{corbeil2025mediqaoe}. The DAU oracle is given the gold order list.

Clarification helps whenever transcript content has been stripped, paraphrased, or left implicit. Table \ref{tab:orders} reports the MEDIQA-OE official Match-F1 metric. DAU improved performance on the degradations that strip explicit order specifics: \textit{Dropped Content}, \textit{Implied Context}, and \textit{Patient-facing}.

\begin{table*}
\centering\small
\setlength{\tabcolsep}{3pt}
% Real col range 44-49; ACI cols mean=74.6 max=87.4; PriMock cols mean=79.3 max=83.9
\newheatcol{q}{min=44, max=49, color=cyan}
\newheatcol{a}{min=70, max=87.4, color=cyan}
\newheatcol{p}{min=75, max=84, color=cyan}
\resizebox{\textwidth}{!}{%
\begin{tabular}{@{}l|q|aaaaaa|pppppp@{}}
\toprule
 & \textbf{\CN} & \multicolumn{6}{c|}{\textbf{ACI-Bench}} & \multicolumn{6}{c}{\textbf{PriMock57}} \\
\textbf{Method} & --- & Original & ASR & Drop & Impl & PatF & Vis & Original & ASR & Drop & Impl & PatF & Vis \\
\midrule
zero\_shot\_yim\_opus46 & 44.2 & 85.7 & 79.6 & 58.3 & 65.4 & 66.2 & 74.9 & 80.6 & 75.9 & 71.6 & 73.3 & 76.4 & 77.5 \\
zero\_shot\_abacha\_opus46 & 45.9 & 86.7 & 80.6 & 59.4 & 66.4 & 66.5 & 77.5 & 81.5 & 78.0 & 73.6 & 74.3 & 78.4 & 78.7 \\
\midrule
DAU Budget=0 & 46.9 & 87.3 & 81.8 & 61.3 & 67.7 & 68.0 & 78.2 & 83.1 & 80.0 & 75.9 & 76.7 & 80.6 & 81.6 \\
DAU Budget=1 & 48.2 & 87.3 & 82.0 & 63.5 & 70.5 & 70.4 & 79.8 & 83.9 & 80.4 & 77.2 & 78.8 & 81.6 & 82.2 \\
DAU Budget=3 & 48.7 & 87.4 & 82.2 & 65.2 & 73.6 & 71.7 & 80.3 & 83.9 & 80.7 & 77.8 & 80.0 & 81.6 & 82.6 \\
DAU Budget=5 & 48.8$^{\ddagger}$ & 87.4$^{\ddagger}$ & 82.3$^{\ddagger}$ & 65.9$^{\ddagger}$ & 74.0$^{\ddagger}$ & 72.0$^{\ddagger}$ & 80.3$^{\ddagger}$ & 83.9$^{\ddagger}$ & 80.7$^{\ddagger}$ & 77.9$^{\ddagger}$ & 80.1$^{\ddagger}$ & 81.6$^{\ddagger}$ & 82.6$^{\ddagger}$ \\
\midrule
DAU Budget=$\infty$ & 97.4 & 99.9 & 99.9 & 99.8 & 99.8 & 99.8 & 99.8 & 97.8 & 97.7 & 97.6 & 97.9 & 97.9 & 97.5 \\
\bottomrule
\end{tabular}
}
\caption{Note completeness (fact-recall, higher is better). Completeness measures fact recall from the transcript against the reference note; it is upper-bounded by what was actually spoken, so scores well below 100\% typically reflect missing chart context or unspoken findings rather than note-generation errors, as evidenced by the similar Yim and Abacha baseline scores.
Real = real ASR transcripts, ($N{=}3{,}000$). Public-dataset columns: Original, ASR, Dropped Content, Implied Context, Patient-facing, Visual/Physical. Significance: per-encounter Wilcoxon signed-rank test of DAU Budget=5 vs.\ the stronger of the two zero-shot baselines. $\dagger\,p{<}0.05$, $\ddagger\,p{<}0.01$.}

\label{tab:notes}
\end{table*}

\begin{table*}
\centering\small
\setlength{\tabcolsep}{3pt}
% Real range 20-45; ACI-note mean=35.5 max=48.2; SynAll mean=31.9 max=69.8; SynMim mean=58.1 max=93.1; MIMIC mean=47.7 max=58.1
\newheatcol{q}{min=20, max=45, color=cyan}
\newheatcol{x}{min=30, max=48.2, color=cyan}
\newheatcol{y}{min=27, max=69.8, color=cyan}
\newheatcol{z}{min=53, max=93.1, color=cyan}
\newheatcol{w}{min=42, max=58.1, color=cyan}
\begin{tabular}{l|q|xxxxxx|yz|w}
\toprule
 & \textbf{\CN} & \multicolumn{6}{c|}{\textbf{ACI-Bench} (note)} & \multicolumn{2}{c|}{\textbf{ICD-Synth}} & \textbf{MIMIC-IV} \\
\textbf{Method} & --- & Original & ASR & Drop & Impl & PatF & Vis & All & MIMIC & Discharge \\
\midrule
two\_step \cite{boyle2023automatedcoding} & 32.0 & 44.9 & 41.1 & 40.2 & 35.9 & 39.6 & 36.1 & 42.6 & 81.1 & 42.8 \\
Retrieval-coder \cite{yang2022kept} & 20.7 & 35.6 & 34.5 & 32.9 & 32.2 & 32.4 & 31.8 & 69.8 & 93.1 & 36.9 \\
%Generate-Expand-Verify \cite{yuan2025reliable} & 32.7 & 33.8 & 31.9 & 31.7 & 32.6 & 29.9 & 31.6 & 63.3 & 89.8 & 42.4 \\
\midrule
CAML \cite{mullenbach2018caml}            & --- & 21.2 & 20.9 & 19.0 & 17.6 & 17.0 & 20.6 & 0.8 & 8.0 & 52.5 \\
LAAT \cite{vu2020laat}                    & --- & 32.2 & 31.5 & 32.1 & 29.6 & 28.3 & 32.5 & 0.8 & 11.7 & 58.1 \\
MultiResCNN \cite{li2020multirescnn}      & --- & 20.9 & 20.1 & 19.2 & 19.7 & 19.5 & 20.6 & 0.6 & 4.0 & 55.3 \\
\midrule
DAU Budget=0 & 31.4 & 43.9 & 41.0 & 39.9 & 36.8 & 37.8 & 36.1 & 43.8 & 81.1 & 41.9 \\
DAU Budget=1 & 38.7 & 45.1 & 42.4 & 42.3 & 43.1 & 45.9 & 41.3 & 42.5 & 81.1 & 46.0 \\
DAU Budget=3 & 43.0 & 47.3 & 45.3 & 46.5 & 45.4 & 47.7 & 44.5 & 42.8 & 81.0 & 47.8 \\
DAU Budget=5 & 44.3$^{\ddagger}$ & 48.2 & 47.7$^{\dagger}$ & 48.0$^{\ddagger}$ & 47.5$^{\ddagger}$ & 47.9$^{\ddagger}$ & 45.3$^{\ddagger}$ & 43.6 & 81.6 & 47.6 \\
\midrule
DAU Budget=$\infty$ & 88.2 & 82.6 & 81.2 & 82.3 & 82.1 & 82.6 & 80.5 & 44.0 & 82.0 & 87.1 \\
\bottomrule
\end{tabular}
\caption{ICD-10 coding F1 on the test splits. Real = real ASR transcripts ($N{=}3{,}000$, no degradation applied; supervised MIMIC-trained coders not run). Significance: per-encounter Wilcoxon signed-rank tests of DAU Budget=5 vs.\ the strongest non-DAU baseline on each column. $\dagger\,p{<}0.05$, $\ddagger\,p{<}0.01$ (Original ACI-Bench not significant).}

\label{tab:coding}
\end{table*}%
\begin{table*}
\centering\small
\setlength{\tabcolsep}{4pt}
% MEDIQA-OE mean=62.3 max=71.1
\newheatcol{o}{min=57, max=71.1, color=cyan}
\begin{tabular}{l|oooooo}
\toprule
 & \multicolumn{6}{c}{\textbf{MEDIQA-OE (test, $N{=}100$)}} \\
\textbf{Method} & Original & ASR & Drop & Impl & PatF & Vis \\
\midrule
\textit{Published MEDIQA-OE 2025 baselines} \\
WangLab (GPT-4) \cite{corbeil2025mediqaoe}       & 81.8 & --- & --- & --- & --- & --- \\
silver\_shaw (Gemini 2.5) \cite{corbeil2025mediqaoe} & 76.4 & --- & --- & --- & --- & --- \\
GPT-4o baseline \cite{corbeil2025mediqaoe}       & 63.6 & --- & --- & --- & --- & --- \\ 
\midrule
zero\_shot\_corbeil\_opus46 \cite{corbeil2025mediqaoe} & 70.1 & 67.4 & 51.3 & 54.0 & 52.2 & 71.1 \\
\midrule
DAU Budget=0 & 70.5 & 67.4 & 51.7 & 54.9 & 52.9 & 70.6 \\
DAU Budget=1 & 69.5 & 68.0 & 59.8 & 60.7 & 60.9 & 69.1 \\
DAU Budget=3 & 69.1 & 68.1 & 65.5 & 67.6 & 64.9 & 69.8 \\
DAU Budget=5 & 69.1 & 68.1 & 65.9$^{\ddagger}$ & 69.1$^{\ddagger}$ & 65.2$^{\ddagger}$ & 69.7 \\
\midrule
DAU Budget=$\infty$ & 76.4 & 77.9 & 84.9 & 79.7 & 78.3 & 77.1 \\
\bottomrule
\end{tabular}
\caption{Medical order extraction, MEDIQA-OE Description Match-F1. Clarification recovers most of the lost performance on degradations that strip or paraphrase order specifics: \textit{Dropped Content} ($+14.6$), \textit{Implied Context} ($+15.1$) and \textit{Patient-facing} ($+13.0$). Budget=$\infty$ is an upper bound on the information a clarification can carry. The gap from 100 reflects noise, not missing information. Significance (per-encounter Wilcoxon signed-rank, DAU Budget=5 vs.\ zero\_shot\_corbeil\_opus46). $\dagger\,p{<}0.05$, $\ddagger\,p{<}0.01$.}

\label{tab:orders}
\end{table*}

%\newpage
%% ============================================================
\section{Analysis}
\label{sec:analysis}

The analyses in this section are restricted to the real corpus so that question quality is measured on real visit dialogue rather than on degraded inputs in order to draw stronger conclusions from these results.
After running DAU with budgets $\in\{1,3,5\}$ this yields $21{,}309$ clarification turns ($9{,}088$ Notes, $12{,}221$ ICD-10).

For every clarification turn we record $\Delta_{\text{metric}}$, the change in the task metric caused by applying that turn's answer. 
Averaged across the turns the mean change is $+1.6$ on Notes and $+5.5$ on ICD-10. The asymmetry comes from the metric scales: a single ICD-10 code addition / removal moves F1 by $\sim$10, whereas one extra recovered fact is worth $\sim$1--5 of completeness.

\subsection{What Makes a Good Question?}
\label{sec:analysis-good-q}

\begin{figure*}[ht]
\centering
\includegraphics[width=\linewidth]{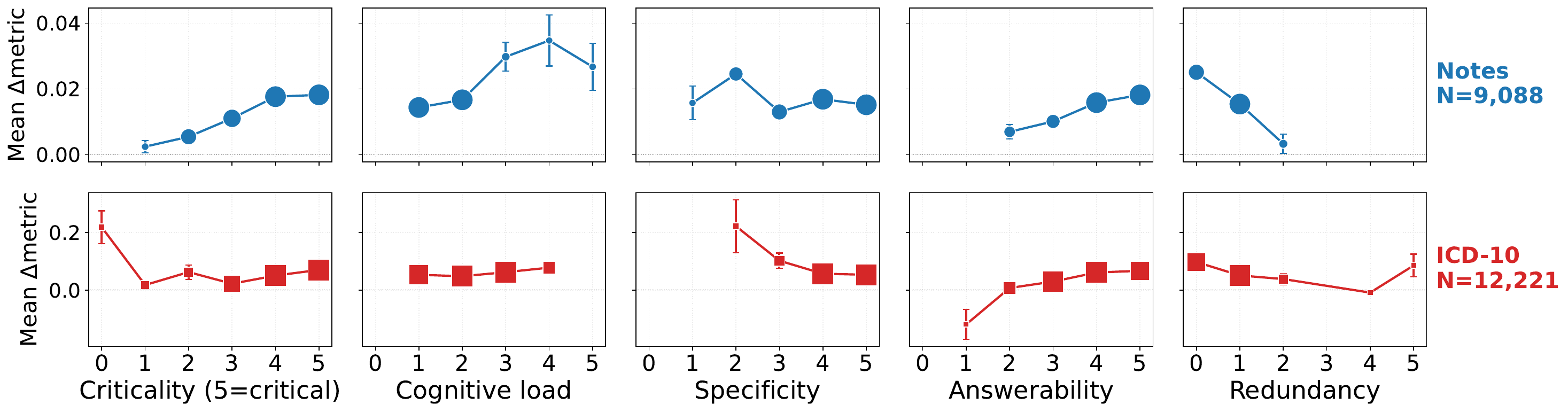}
\caption{Real data only: mean per-turn change in the task metric vs.\ each $0$--$5$ rubric score; top row Notes ($N{=}9{,}088$), bottom row ICD-10 ($N{=}12{,}221$), with mean$\pm$s.e.\ error bars (bins with $<$10 turns dropped). Marker area is proportional to $\sqrt{N_{\text{bin}}}$, so visibly small markers indicate sparse bins. The five rubrics: \textbf{criticality}, how clinically consequential it is to resolve the gap ($5{=}$critical); \textbf{cognitive load}, how hard the question is for the clinician to answer; \textbf{specificity}, how narrowly the question targets a single missing fact; \textbf{answerability}, how readily the question can be answered from the encounter; \textbf{redundancy}, how much the answer is already implied by the current draft ($0{=}$brand-new information).}
\label{fig:score-perf}
\end{figure*}

\begin{figure*}[ht]
\centering
\includegraphics[width=\linewidth]{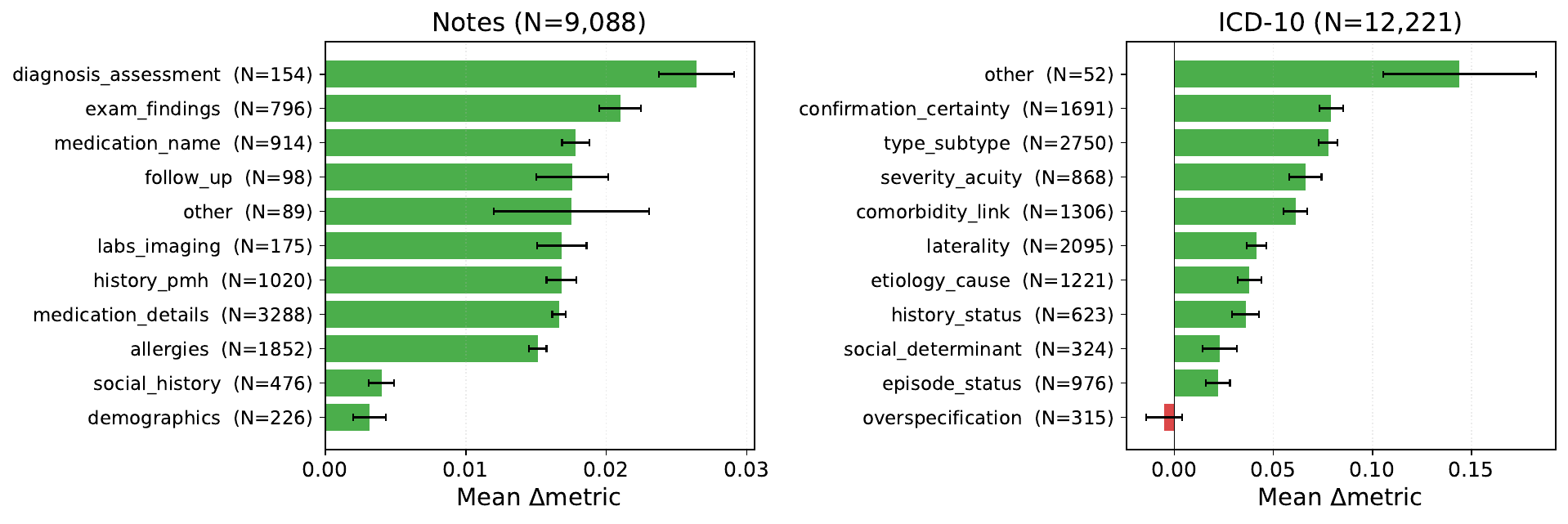}
\caption{Real data only: mean per-turn $\Delta_{\text{metric}}$ by question category, one panel per task (Notes left, ICD-10 right; bar = category mean, error bars = $\pm$s.e., $N$ shown next to each label; categories with $<$20 turns dropped). Categories with negative mean $\Delta$ (where the rewrite hurts the artifact more than it helps) are colored red. The only negative bar on ICD-10 is \emph{overspecification} (questions asking for external-cause / place / activity detail that results in overly specific codes). The encounter rarely documents these, so the answer doesn't add a new code that matches the gold, and the rewrite often perturbs an already-correct code, so F1 goes down.}
\label{fig:goodq}
\end{figure*}

In order to build a system that asks the most informative questions we study which features predict the per-turn change in the task metric ($\Delta_{\text{metric}}$). Every question is scored post-hoc by an LLM judge on five $0$--$5$ rubrics; the rubrics were developed with clinical experts and evaluated with Claude Opus~4.6 (Appendix~\ref{sec:appendix-rubrics}). Figure~\ref{fig:score-perf} plots the mean $\Delta_{\text{metric}}$ inside each rubric bin, and Figure~\ref{fig:goodq} aggregates by question category.

Table~\ref{tab:useful-corr} reports the per-feature Pearson correlation with $\Delta_{\text{metric}}$ across the $21{,}309$ Real data clarification turns.
Oracle confidence is the strongest individual predictor ($r{=}{+}0.164$, $p{<}0.001$): a confident oracle answer is associated with a larger metric improvement. Cognitive load is next ($r{=}{+}0.072$, $p{<}0.001$): harder questions extract more useful information when they land. Redundancy follows ($r{=}{-}0.054$, $p{<}0.001$): questions whose answer was already in the draft move the metric down on average, since any rewrite then risks over-writing correct content. Criticality ($r{=}{+}0.052$, $p{<}0.001$) also correlates positively: higher-stakes questions are associated with larger gains. Turn index ($r{=}{-}0.026$) and answerability ($r{=}{+}0.023$) are weakly significant; specificity has no significant linear effect on this corpus.

\begin{table}[t]
\centering
\small
\begin{tabular}{ll}
\toprule
Feature & Pearson $r$ \\
\midrule
Oracle confidence & $+0.164$\,$^{\ddagger}$ \\
Cognitive load    & $+0.072$\,$^{\ddagger}$ \\
Redundancy        & $-0.054$\,$^{\ddagger}$ \\
Criticality       & $+0.052$\,$^{\ddagger}$ \\
Turn index        & $-0.026$\,$^{\ddagger}$ \\
Answerability     & $+0.023$\,$^{\ddagger}$ \\
Specificity       & $+0.005$\,(n.s.) \\
\bottomrule
\end{tabular}
\caption{Real data only: Pearson correlation of each feature with the per-turn $\Delta_{\text{metric}}$. $N{=}21{,}309$ clarification turns. Significance is the Wald $t$-test, $\dagger\,p{<}0.05$, $\ddagger\,p{<}0.01$; n.s.\ = not significant at $p<0.05$.}
\label{tab:useful-corr}
\end{table}

\subsection{Capping Question Complexity}
\label{sec:analysis-complexity}

\begin{figure}[t]
\centering
\includegraphics[width=\linewidth]{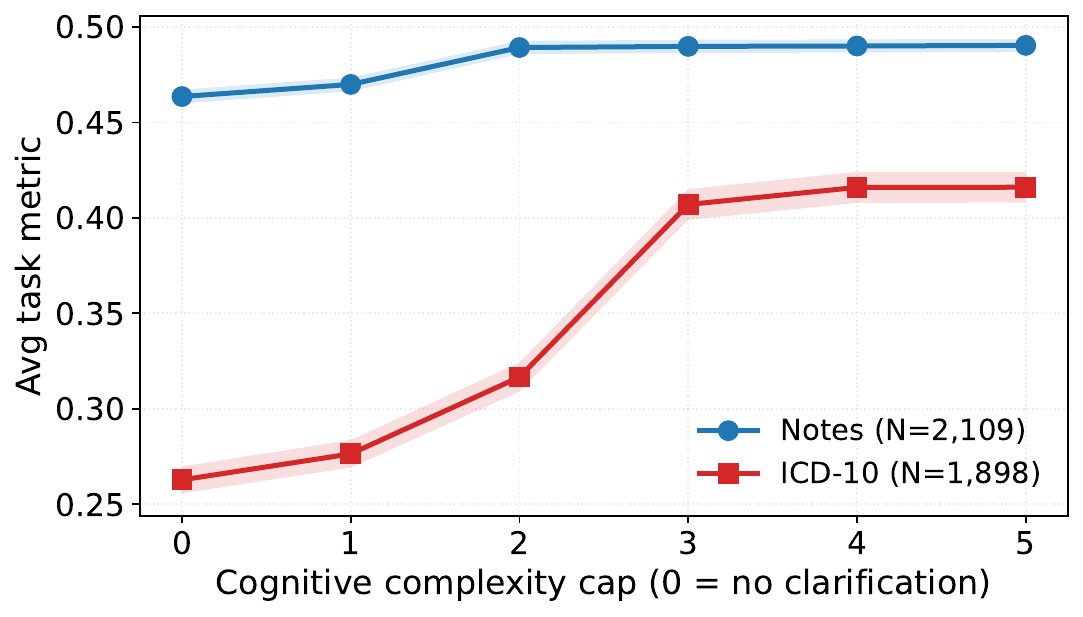}
\caption{Real data only: task performance when the DAU loop is stopped at the first question exceeding a cognitive-load cap ($0{=}$no clarification, $5{=}$allow all), averaged over budget-5 trajectories, shaded bands are $\pm$s.e. Notes saturate with only simple questions (cap 2); ICD-10 needs the harder, multi-option questions to reach full performance.}
\label{fig:complexity}
\end{figure}

If cognitive load is positively associated with $\Delta_{\text{metric}}$ (\S\ref{sec:analysis-good-q}), how much of the gain actually depends on the hard questions? On real data we replay every budget-5 trajectory under a complexity cap: the loop keeps its questions in order but stops at the first question whose cognitive-load score exceeds the threshold. Figure \ref{fig:complexity} plots the resulting task metric against the cap.

Two regimes emerge. Notes saturate at $2$: allowing only trivial yes/no and single-fact recall questions already captures essentially all of the budget-5 gain. ICD-10 keeps improving through $4$, the fine-grained distinctions coding requires (laterality, acuity, structure-level anatomy) are exactly the multi-option questions the rubric scores as complex. The practical implication: for note completeness a system can restrict itself to low-burden questions at almost no cost, but for coding the high-burden questions are where the value is, consistent with the positive cognitive-load correlation in Table \ref{tab:useful-corr}.

\subsection{When Clarification Hurts}
\label{sec:analysis-harmful}
On real data $1{,}834$ of the $21{,}309$ clarification turns ($9\%$) had $\Delta_{\text{metric}}{<}0$: the update lowered the task metric. The negative-$\Delta$ rate is task-specific: $9\%$ on ICD-10 ($1{,}145/12{,}221$) versus $8\%$ on Notes ($689/9{,}088$).
To understand the mechanism we feed the real data negative-$\Delta$ turns to an LLM that induces a question-side and an answer-side taxonomy independently and then cross-labels every turn with one mode from each.

\begin{figure}[t]
\centering
\includegraphics[width=\linewidth]{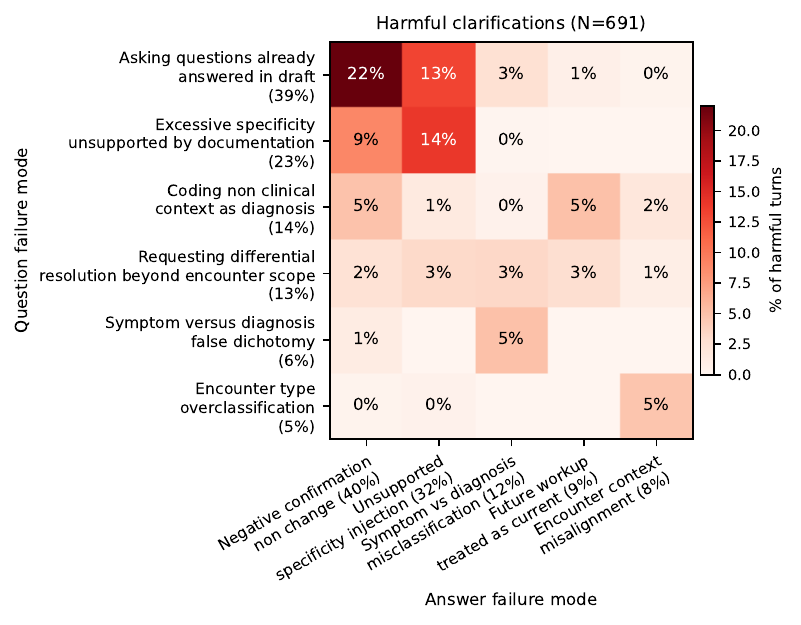}
\caption{Real data only: negative-$\Delta_{\text{metric}}$ clarification turns ($N{=}1{,}834$) broken down by question failure mode (rows) and answer failure mode (columns); cells show the share of all such turns and row/column percentages are appended to the labels. The dominant pattern is redundant questions answered with negative-confirmation non-change.}
\label{fig:harmful}
\end{figure}

Figure \ref{fig:harmful} cross-tabulates the two. The dominant question-side failure is \emph{redundant information request} ($39\%$): the question targets a fact already settled in the draft or transcript, so rewrites introduce errors. Followed by \emph{excessive specificity unsupported by documentation} ($23\%$) and \emph{coding non-clinical context as diagnosis} ($14\%$), questions that demand a degree of detail or a categorical commitment the encounter never makes. 

The dominant answer-side failure is \emph{negative confirmation non-change} ($40\%$): the oracle answer says no new information, yet the artifact is still rewritten. \emph{Unsupported specificity injection} ($32\%$) is next: the oracle answer adds laterality, mechanism, dosage, or subtype that the source does not actually document. The single largest cell ($22\%$ of harm) is redundant questions answered with negative-confirmation non-change, the loop revisits a settled detail and rewrites the artifact even though the oracle confirmed nothing new. 

The two avoidance rules this implies are: (i) do not edit the draft when the answer is a non-answer, and (ii) do not re-ask or adopt specificity for facts already settled in the current draft.

\section{Conclusion}

In this work we explore how to automate provider queries to improve performance on three clinical tasks and propose a straightforward iterative loop DAU (Draft, Ask, Update) that transfers across these domains.
We explore real errors on a difficult-to-obtain real world dataset of transcribed conversations and final clinical notes.
These insights are used to create a dataset of degraded transcripts to test five different degradation categories mimicking real life failure modes.
On these degraded transcripts we can see DAU recovers most of the lost performance specifically on degradations that strip or paraphrase content (Dropped Content, Implied Context, and Patient-facing), while Original and ASR inputs barely move which is consistent with our real world evidence.

We analyze the questions and answers from the DAU loop on real world data to learn what makes good and bad questions. (i) When it helps: gains are concentrated on transcripts with missing content (Dropped Content, Implied Context). (ii) What predicts a metric-improving question: oracle answer confidence is by far the strongest positive signal and redundancy the strongest negative signal. (iii) How hard the questions need to be is task-specific: on real data, capping question complexity shows note completeness retains nearly all of its gain with only simple recall questions (cap 2), while ICD-10 coding needs the harder multi-option questions through cap 4 to reach full F1. (iv) Why it sometimes hurts: $\sim$$8$--$10\%$ of turns drive the metric down, concentrated in ICD-10 and dominated by redundant questions whose answers introduce unsupported specificity, plus non-answers that nonetheless trigger a rewrite.

These findings suggest that the next gains in automated provider queries will come less from asking more questions and more from learning when not to ask and when not to rewrite, with oracle confidence, redundancy, and answer-mode signals serving as cheap gates on the loop.

\section*{Limitations}

\paragraph{The oracle is an upper bound.}
The simulated clinician sees the original transcript and the gold reference, so our results measure the ceiling of what clarification can recover, not performance with a real provider, who may be mistaken or unwilling to answer.

\paragraph{LLM-as-everything.}
One backbone (Claude Opus 4.6) serves as drafter, question generator, oracle, rubric judge, and evaluator, and the degradations are LLM-generated (DeepSeek V3.2); none of these judgments are human-validated, so self-preference effects could inflate results.

\paragraph{Clinician burden is not modeled directly.}
We approximate clinician burden by capping answer length at five words but do not measure the workflow cost of a provider answering questions (time, alert fatigue, willingness to respond), a known friction point for human CDI queries \cite{rouse2022cdi}.

\paragraph{Motivating-audit categories are LLM-assigned.}
The cause categories in Table~\ref{tab:cohort-gap-intro} were assigned by the same LLM family used elsewhere in the paper, on a single real data sample with EHR-derived gold notes that contain chart-pulled history and PHI-randomized identifiers; the percentages should be read as a directional decomposition rather than a validated taxonomy. We did not have human coders re-label the missing facts.

\section*{Data availability}
The Real corpus is internal and not redistributable. ACI-Bench (CC-BY-4.0), PriMock57 (CDLA-Permissive-2.0), and MEDIQA-OE 2025 (CC-BY-4.0) are publicly available. MIMIC-IV-Note is available to credentialed users via PhysioNet. The five degraded-transcript benchmark datasets generated in this work will be released at the address \url{TO-BE-DETERMINED}.

\section*{Code availability}
The code used in this study is available at \url{TO-BE-DETERMINED}.

\section*{Acknowledgements}
This work was funded by Amazon Inc. All authors performed this work while employed by Amazon Inc.

\section*{Author contributions}
J.P.C., R.S., J.M., and H.C.S. conceived of the study. F.W., C.S., and S.N. advised.

\section*{Competing interests}
All authors are or were employees of Amazon Inc during the course of this work. The authors declare no other financial or non-financial competing interests.

\bibliography{custom}

\clearpage

\appendix

\section{Datasets}
\label{sec:appendix-datasets}

Test-split sizes per dataset and task are shown in Table~\ref{tab:dataset-summary}. We always use the test split.

\begin{table}[h]\small
\centering
\begin{tabular}{lccc}
\toprule
Dataset & Notes & ICD-10 & Orders \\
\midrule
Real       & 3{,}000 & 3{,}000 & --- \\
ACI-Bench                 & 120 & 120 & --- \\
PriMock57                 & 57 & --- & --- \\
MIMIC-IV discharge        & --- & 8{,}066 & --- \\
ICD-Synth (this work)     & --- & 880\,/\,714 & --- \\
MEDIQA-OE                 & --- & --- & 100 \\
\bottomrule
\end{tabular}
\caption{Test-split sizes. Public datasets are also run on the five 20\% degradations of \S\ref{sec:degradations} at the same $N$.}
\label{tab:dataset-summary}
\end{table}

\paragraph{Real data (internal).} 3{,}000 primary-care encounters from US hospitals (2025), private and internal. Each has an ASR transcript and an EHR-derived clinician SOAP note with billed ICD-10-CM codes. All data is anonymized. Real is internal and not redistributable.

\paragraph{ACI-Bench \citep{yim2023acibench}.} Public synthesized doctor-patient SOAP corpus (207 encounters; we use the 120-encounter test split). For ICD-10 we use the expert codes from \citet{yuan2025reliable}. CC-BY-4.0.

\paragraph{PriMock57 \citep{korfiatis2022primock57}.} 57 mock primary-care consults with audio, transcript, and clinician note. Used only for note completeness. CDLA-Permissive-2.0.

\paragraph{MIMIC-IV \citep{johnson2023mimiciv} discharge notes.} Discharge-summary subset of MIMIC-IV-Note (PhysioNet, credentialed). Gold ICD-10 codes from the billing table. Used only for ICD-10. The CAML / LAAT / MultiResCNN baselines are the \citet{edin2023automated} checkpoints (trained on this train split).

\paragraph{ICD-Synth (this work).} Synthetic ICD-10 sanity check. We sample $k\sim\text{Unif}\{1,\dots,5\}$ ICD-10-CM codes and concatenate their 2025 long descriptions. \textbf{All} ($\sim$880) draws from the full ICD-10-CM vocabulary; \textbf{MIMIC} ($\sim$714) restricts to codes in the MIMIC-IV train split.

\paragraph{MEDIQA-OE 2025 \citep{corbeil2025mediqaoe}.} Public shared-task corpus for order extraction (medications, labs, imaging, follow-ups) built on ACI-Bench. Test split $N{=}100$; official Description Match-F1. CC-BY-4.0.

\paragraph{Degradations.} We create a degraded dataset from ACI-Bench by applying each of the five 20\% degradations of \S\ref{sec:degradations} (ASR, Dropped, Implied, Patient-facing, Visual/Physical) independently to the test transcripts. Same $N$ as the original split, so per-column Wilcoxon comparisons are paired.

\section{Real Provider Queries}
\label{appendix:real-queries}

We interviewed a clinical scribe to learn how provider queries were used and to identify examples which motivated the question generation prompts and analysis: 

\begin{enumerate}[noitemsep,topsep=2pt]
    \item What are the patient's final diagnoses?
    \item Can you confirm the patient's current medications?
    \item What were the pertinent negative and positive physical exam findings?
    \item What is the patient's complete physical exam?
    \item What are the patient's chronic medical conditions?
    \item For what condition is the patient being prescribed depo-testosterone?
    \item Is schizophrenia the cause of the patient's tremors, or does the patient have another condition causing them?
    \item What was the medication name of the biweekly injections the patient is receiving?
    \item What were the diagnoses/assessment items for the visit?
    \item What are the specific types/locations of the patient's abdominal hernia(s)?
    \item What psychiatric condition(s) was the patient diagnosed with prior to visit and are they the same as final diagnoses for the visit?
    \item What medications did outside provider prescribe for stomach pain and nausea?
    \item What medication was patient previously taking for toe pain?
    \item What was heard on lung auscultation?
\end{enumerate}

\section{Prompt Templates}
\label{sec:appendix-prompts}
We include the exact system and user prompts for the two LLM calls inside the DAU loop, the question generator and the oracle, for each of the three tasks. Curly-braced tokens are Python format-string placeholders filled in per turn (transcript, current draft, previous Q\&A, gold reference, etc.). The note-generation prompt and downstream evaluators are not LLM-task-specific to DAU and are reproduced in the source repository.

\subsection{Clinical Notes}

\begin{promptbox}{Notes -- Question generator (system)}
\begin{Verbatim}
You are a clinical quality reviewer auditing a medical scribe's draft note. Your job is to ask the SINGLE clarifying question that, if answered, would most improve the completeness of the note , i.e. most likely surface clinically relevant content that is MISSING from the current draft. Never assume a specific finding, drug, or value exists unless the transcript or draft note already mentions it. If the draft already adequately covers the transcript and no clarifying question would meaningfully improve completeness, you MUST abstain instead of inventing a question.
\end{Verbatim}
\end{promptbox}

\begin{promptbox}{Notes -- Question generator (user template)}
\begin{Verbatim}
Based on the transcript and the current draft note, ask the ONE clarifying question that would most improve note completeness. It must target a clinically relevant fact MISSING from the draft and most likely to add a gold item.

CLINICAL TRANSCRIPT:
{transcript_text}

CURRENT DRAFT NOTE:
{note_section}
{uncovered_hint}
{prev_section}
{question_limit}{word_limit_section}
HOW TO CHOOSE THE QUESTION (broad -> narrow):
1. If a core note section is missing entirely, ask a broad question about it.
2. Once a broad area is covered, drill into dosages, durations, laterality, etc.
3. Never invent specific drug/test/value names not in the transcript or draft.
4. Do NOT repeat or re-ask a question already in PREVIOUS CLARIFICATIONS.
5. Prefer the gap whose resolution would most improve completeness right now.

ABSTENTION:
If the current draft already adequately covers the transcript, OR no clarifying
question would meaningfully improve note completeness, return:
{"question": null, "abstained": true, "reason": "<short reason>"}

OUTPUT FORMAT:
Return ONLY a JSON object. Either:
  {"question": "<your question>", "abstained": false}
or:
  {"question": null, "abstained": true, "reason": "<short reason>"}
\end{Verbatim}
\end{promptbox}

\begin{promptbox}{Notes -- Oracle (system)}
\begin{Verbatim}
You are a healthcare provider who was present for this clinical encounter. You have access to the full dialogue transcript AND the authoritative documentation facts from the encounter. Answer clarifying questions factually based on both sources.
\end{Verbatim}
\end{promptbox}

\begin{promptbox}{Notes -- Oracle (user template)}
\begin{Verbatim}
You have two sources of truth about this encounter:

ENCOUNTER TRANSCRIPT:
{transcript_block}

AUTHORITATIVE DOCUMENTATION FACTS (from the finalized clinical note):
{facts_block}

CLARIFYING QUESTION FROM THE SCRIBE:
{question}

CRITICAL , ANSWER LENGTH LIMIT:
Your answer MUST be {max_answer_words} words or fewer. Use only essential
clinical terms. Examples:
- "Lisinopril 40mg daily for hypertension" (5 words)
- "BP 142/89, HR 78" (4 words)
- "Return in 2 weeks" (4 words)
- "No imaging ordered" (3 words)
- "No information on this" (4 words, for unknown)

Respond in JSON:
{
  "answer": "Your concise clinical answer",
  "confidence": 0.0 to 1.0
}

GUIDELINES:
- Prefer information from the AUTHORITATIVE FACTS when it conflicts with the
  transcript , the facts are the canonical documentation.
- Supplement with transcript details (e.g. verbatim quotes, timing, nuance)
  when the question asks for them and they are not in the facts.
- If NEITHER source contains the information, answer exactly "No information
  on this" (or within the word limit) , do not invent content.
- Be specific: include numbers, units, dosages, frequencies, anatomic sides.
- confidence: 1.0 = directly stated in the facts, 0.7 = stated in the
  transcript only, 0.4 = inferred, 0.0 = not present.
\end{Verbatim}
\end{promptbox}

\subsection{ICD-10 Coding}

\begin{promptbox}{ICD-10 -- Question generator (system)}
\begin{Verbatim}
You are a senior medical coder reviewing another coder's work. Ask the SINGLE most important clarifying question to fix a coding error , either a missing diagnosis OR insufficient specificity , so that the final ICD-10 code set most improves. Never mention specific ICD-10 code strings; ask about clinical facts. If the current code set already adequately captures the clinical text and no clarifying question would meaningfully improve coding F1, you MUST abstain instead of inventing a question.
\end{Verbatim}
\end{promptbox}

\begin{promptbox}{ICD-10 -- Question generator (user template)}
\begin{Verbatim}
You are auditing a draft set of ICD-10 codes. Ask the ONE clarifying question that would most improve coding accuracy by closing the single most important coding gap.

CLINICAL TEXT:
{clinical_text}

CURRENT CODES + DESCRIPTIONS:
{code_block}
{prev_block}

CODING AXES TO CONSIDER:
{axes_block}
{q_limit}{word_constraint}
HARD RULES:
- Never mention an ICD-10 code string in the question.
- Ask about exactly ONE coding gap.
- Do NOT repeat a question already in PREVIOUS Q&A.
- Prefer the gap whose resolution would most improve coding F1 right now.

ABSTENTION:
If the current codes already adequately cover the clinical text, OR no
clarifying question would meaningfully improve coding F1, return:
{"question": null, "abstained": true, "reason": "<short reason>"}

OUTPUT FORMAT:
Return ONLY a JSON object. Either:
  {"question": "<your question>", "abstained": false}
or:
  {"question": null, "abstained": true, "reason": "<short reason>"}
\end{Verbatim}
\end{promptbox}

\noindent\textit{Coding axes injected via} \texttt{\{axes\_block\}}\textit{:} active diagnoses; chronic comorbidities; acute vs chronic; laterality (right / left / bilateral); severity / stage; type / subtype; initial vs subsequent encounter; complications; pregnancy / trimester (if applicable); causal / underlying disease relationships.

\begin{promptbox}{ICD-10 -- Oracle (system)}
\begin{Verbatim}
You are the documenting clinician from this encounter. You have the full original clinical text AND the authoritative list of diagnoses that were billed for this encounter. Answer clarifying questions from a medical coder factually, grounded in what was actually documented. You MUST NOT mention specific ICD-10 codes , only describe the underlying clinical facts (diagnoses, severity, laterality, acuity, comorbidities).
\end{Verbatim}
\end{promptbox}

\begin{promptbox}{ICD-10 -- Oracle (user template)}
\begin{Verbatim}
A medical coder is trying to assign ICD-10 codes for this encounter. They have two sources of truth about the encounter:

ORIGINAL CLINICAL TEXT:
{original_text}

AUTHORITATIVE CONFIRMED DIAGNOSES (what was billed for this encounter):
{dx_block}

CODER'S QUESTION:
{question}

HARD LIMIT: Answer in {max_answer_words} words or fewer.

Respond in JSON:
{"answer": "...", "confidence": 0.0 to 1.0}

RULES:
- Never mention ICD-10 code strings (e.g. "E11.9", "I10"). Describe the
  clinical condition in words only.
- Prefer information from the AUTHORITATIVE DIAGNOSES when the transcript
  is ambiguous.
- Be specific: include severity, laterality, acuity, type, stage if they
  were documented.
- If the question asks about something NOT documented, answer exactly
  "Not documented" , do not speculate.
- confidence: 1.0 = explicit in diagnoses, 0.7 = clear in transcript,
  0.4 = inferred, 0.0 = absent.
\end{Verbatim}
\end{promptbox}

\noindent\textit{In addition, the oracle's free-text answer is post-processed by a regex} \verb|[A-TV-Z][0-9][0-9A-Z]|\allowbreak\verb|(?:\.[0-9A-Z]{1,4})?| \textit{that replaces any ICD-10-shaped token with} \texttt{[CODE\_REDACTED]} \textit{so the oracle cannot leak the gold codes verbatim.}

\subsection{Medical Order Extraction}

\begin{promptbox}{Orders -- Question generator (system)}
\begin{Verbatim}
You are a clinical quality reviewer verifying medical orders from doctor-patient transcripts. Ask the SINGLE clarifying question that would most improve the order list -- either by uncovering an order MISSING from the current extraction or by RESOLVING a vague/lay-language order into a specific clinical one.

An order is ADEQUATE only when it has the specifics a clinical record requires:
  - medication: drug name, dose, and frequency/route
  - lab: specific test/panel name (not just 'labs' or 'bloodwork')
  - imaging: modality AND body region/structure (not 'a scan' or 'lower area')
  - followup: timing and the clinician/specialty (or condition trigger)

Treat lay paraphrases ('water pill', 'stronger pain medicine', 'something for the inflammation', 'a scan of the lower area', 'some tests') and bare type mentions ('some labs', 'a follow-up') as INCOMPLETE -- they need a specifying clarification. Never invent an order that the transcript does not imply. You may ABSTAIN only when (a) every transcript-implied order type already has at least one entry AND (b) every current entry already meets the specificity criteria above.
\end{Verbatim}
\end{promptbox}

\begin{promptbox}{Orders -- Question generator (user template)}
\begin{Verbatim}
Based on the transcript and currently extracted orders, ask the ONE clarifying question that would most improve the order list by recovering a MISSING order or by RESOLVING a vague/lay-language order into a specific clinical one.

CLINICAL TRANSCRIPT:
{transcript_text}
{orders_section}
{uncovered_hint}{vagueness_hint}
{prev_section}
{question_limit}{word_limit_section}
HOW TO CHOOSE THE QUESTION (broad -> narrow):
1a. If an order type has NO entries yet, ask a broad question about it.
1b. If a current order is VAGUE (lay paraphrase, missing drug/dose/frequency,
    missing test/panel name, missing imaging modality+region, missing follow-up
    timing), ask a SPECIFYING question to fill in that detail. Vague orders are
    just as much a gap as missing ones.
2. Once a broad area is covered with specifics, refine remaining details.
3. Do NOT invent specific drug/test/imaging names not in the transcript -- but
   you MAY ask the provider to name the specific drug/test/modality.
4. Do NOT repeat a question already in PREVIOUS CLARIFICATIONS.
5. Prefer the gap (missing OR vague) whose resolution would most improve the
   order list right now.

ABSTENTION:
Abstain ONLY if (a) every transcript-implied order type already has an entry
AND (b) every current order entry is already specific (drug name+dose+freq, or
specific test/panel, or modality+region, or follow-up timing+clinician). If any
listed order is still a lay paraphrase or generic placeholder, do NOT abstain.
{"question": null, "abstained": true, "reason": "<short reason>"}

OUTPUT FORMAT:
Return ONLY a JSON object. Either:
  {"question": "<your question>", "abstained": false}
or:
  {"question": null, "abstained": true, "reason": "<short reason>"}
\end{Verbatim}
\end{promptbox}

\noindent\textit{Auto-injected hints:} \texttt{\{uncovered\_hint\}} lists order types (medication, lab, imaging, followup) that have no entry yet; \texttt{\{vagueness\_hint\}} runs a regex heuristic over each current order's description and flags lay-language phrases (\textit{some labs, water pill, stronger pain medicine, a scan, lower area, follow-up visit}, etc.), medications missing a dose/frequency number, labs/imaging missing a specific test/modality, and follow-ups missing timing or specialty. Both are passed back to the LLM as bullet lists so the question targets a real gap rather than re-asking already-specified orders.

\begin{promptbox}{Orders -- Oracle (system)}
\begin{Verbatim}
You are a healthcare provider who was present for this clinical encounter. You have access to the complete list of medical orders placed during the visit. Answer questions about what was ordered, dosages, reasons, and follow-up plans.
\end{Verbatim}
\end{promptbox}

\begin{promptbox}{Orders -- Oracle (user template)}
\begin{Verbatim}
You have access to the following ground-truth medical orders from this encounter:

MEDICAL ORDERS:
{orders_text}

ENCOUNTER TRANSCRIPT (for context):
{transcript_text}

CLARIFYING QUESTION:
{question}

CRITICAL , ANSWER LENGTH LIMIT:
Your answer MUST be {max_answer_words} words or fewer. Be concise , use only essential
clinical terms. Examples:
- "Lasix 40mg daily" (3 words)
- "Hemoglobin A1c, CBC" (3 words)
- "MRI left shoulder" (3 words)
- "Yes, pregnancy test ordered" (4 words)
- "No imaging ordered" (3 words)

Respond in JSON:
{
  "answer": "Your clinical answer about the orders",
  "confidence": 0.0 to 1.0
}

GUIDELINES:
- Answer based ONLY on the orders listed above
- If asked about an order type that has no entries, say "No [type] orders were placed"
- Be specific: include medication names, dosages, lab test names, imaging modalities
- confidence: 1.0 = directly in the order list, 0.5 = inferred from context, 0.2 = best guess
- Keep answers concise and clinically relevant
\end{Verbatim}
\end{promptbox}

\section{Question and Answer Length}
\label{sec:appendix-length}
The generator is capped at 20 words per question and the oracle at 5 words per answer. Actual usage across the 12,487 clarification turns in our budget-5 sweep (Table \ref{tab:qalength}) shows the generator self-regulates, averaging 9.7 words on Notes, 15.0 on ICD-10, and 7.6 on Orders (roughly half the cap). The oracle, in contrast, almost always saturates the answer budget: the mean is 3.8--4.2 words out of 5, and the median is exactly 4 words in every domain. ICD-10 questions are the longest (medically-specific phrasing like "Was the dysphagia resolved or is it chronic?"); order-extraction questions are the shortest (often binary yes/no about a specific drug or lab).

\begin{table}[ht]
\centering\small
\begin{tabular}{lrrrrr}
\toprule
Domain & $N_q$ & $\bar{Q}_w$ & $\tilde{Q}_w$ & $\bar{A}_w$ & $\tilde{A}_w$ \\
\midrule
Notes    & 4,099 &  9.7 &  9 & 3.94 & 4 \\
ICD-10   & 6,789 & 15.0 & 15 & 3.80 & 4 \\
Orders   & 1,599 &  7.6 &  8 & 4.24 & 4 \\
\bottomrule
\end{tabular}
\caption{Question ($Q$) and answer ($A$) lengths in words, averaged over every clarification turn of every budget-5 run. Question cap $=$ 20 words, answer cap $=$ 5 words. $N_q$ is the number of clarification turns contributing to the averages.}
\label{tab:qalength}
\end{table}

\newpage
\section{Question Scoring Rubrics}
\label{sec:appendix-rubrics}
Each clarification turn is scored post-hoc by an LLM judge (Claude Opus~4.6) on five $0$--$5$ rubrics. The rubrics were drafted with clinical input and are designed to be mutually independent: \textbf{criticality} (consequence of the gap), \textbf{cognitive load} (burden on the answerer), \textbf{specificity} (how narrowly scoped), \textbf{answerability} (is the fact in the source), and \textbf{redundancy} (already documented?). The full anchor definitions used in the judge prompt are reproduced below; one task-specific example accompanies each level.

\newtcolorbox{rubricbox}[1]{
  enhanced, breakable, colback=blue!2, colframe=blue!40!black,
  colbacktitle=blue!40!black, coltitle=white,
  fonttitle=\bfseries\small\sffamily,
  title={#1},
  boxrule=0.4pt, arc=1.5pt, left=5pt, right=5pt, top=3pt, bottom=3pt,
  before skip=6pt, after skip=6pt,
}
\newcommand{\rlevel}[2]{\item[\textbf{#1}]\textit{#2}\hfill\\}

\begin{rubricbox}{Criticality \,---\, consequence of leaving the gap unresolved}
\footnotesize
Measures how much the correctness/safety of the final artifact (note, ICD-10 codes, or order list) is harmed if the question goes unanswered. Independent of how hard the question is to answer or how it is phrased.
\begin{description}[leftmargin=1.2em,itemsep=1pt]
  \rlevel{5 \,---\, Critical / safety}{Patient harm or a materially wrong artifact, e.g.\ a missing active diagnosis, wrong laterality, or an entirely missing medication order.}
  \rlevel{4 \,---\, Major}{Changes correctness but not safety: a key exam finding, acute vs.\ chronic, wrong dose / frequency.}
  \rlevel{3 \,---\, Moderate}{Improves accuracy or specificity; unlikely to flip the top-line conclusion (severity, subtype, indication).}
  \rlevel{2 \,---\, Minor}{Completeness only: social history, a non-contributory comorbidity, follow-up minutiae.}
  \rlevel{1 \,---\, Trivial}{Cosmetic / near-redundant with information already present.}
  \rlevel{0 \,---\, Negligible}{Off-task or purely exploratory; no realistic effect on the artifact.}
\end{description}
\end{rubricbox}

\begin{rubricbox}{Cognitive load \,---\, burden on the answering clinician}
\footnotesize
Driven by four dimensions and scored as the dominant demand, not a sum: (i)~recall specificity, (ii)~reasoning vs.\ lookup, (iii)~number of options to discriminate among, (iv)~compound / multi-part structure.
\begin{description}[leftmargin=1.2em,itemsep=1pt]
  \rlevel{5 \,---\, Very complex}{Compound multi-part question or a determination integrating many facts. \emph{``Is the heart failure systolic, diastolic, or combined, and what is the EF?''}}
  \rlevel{4 \,---\, Complex}{Synthesis of multiple facts or a fine-grained ($\geq$3-way) technical distinction. \emph{``Medial meniscus tear of anterior horn, posterior horn, or body?''}}
  \rlevel{3 \,---\, Moderate}{Recall plus light judgment, or a 2--3-way choice. \emph{``Single episode or recurrent depression?''}}
  \rlevel{2 \,---\, Simple recall}{One stored fact: a dose, date, laterality, or name. \emph{``What is the dose of Norvasc?''}}
  \rlevel{1 \,---\, Trivial}{Binary / immediate yes-no. \emph{``Should a follow-up be scheduled?''}}
  \rlevel{0 \,---\, None}{No real question is asked, or the answer is stated in the question.}
\end{description}
\end{rubricbox}

\begin{rubricbox}{Specificity \,---\, how narrowly the question is scoped}
\footnotesize
Scope of the \emph{ask}, independent of how important or answerable the fact is. Long wording does not lower specificity.
\begin{description}[leftmargin=1.2em,itemsep=1pt]
  \rlevel{5 \,---\, Pinpointed}{One exact attribute with a single correct answer. \emph{``What was the left knee ROM in degrees?''}}
  \rlevel{4 \,---\, Well-scoped}{A single well-scoped fact with little ambiguity. \emph{``Which lobe was removed during the lobectomy?''}}
  \rlevel{3 \,---\, Focused topic}{A focused topic that may span a few possible facts. \emph{``What were the cardiac exam findings?''}}
  \rlevel{2 \,---\, Broad}{A broad topic with many possible answers. \emph{``Are there any other symptoms?''}}
  \rlevel{1 \,---\, Open-ended sweep}{Vague open sweep over the whole encounter. \emph{``Anything else important?''}}
  \rlevel{0 \,---\, No target}{No real informational target.}
\end{description}
\end{rubricbox}

\begin{rubricbox}{Answerability \,---\, is the fact contained in the source}
\footnotesize
Whether the kind of fact being asked is the kind the encounter actually contains, \emph{not} whether a clinician could guess it.
\begin{description}[leftmargin=1.2em,itemsep=1pt]
  \rlevel{5 \,---\, Clearly contained}{The transcript clearly contains the answer. \emph{``Which medication was the patient told to continue?''}}
  \rlevel{4 \,---\, Likely recoverable}{Recoverable with no real inference. \emph{``What was the chief complaint?''}}
  \rlevel{3 \,---\, Plausible with mild inference}{Mild inference from what was said. \emph{``Is the cough productive or dry?''}}
  \rlevel{2 \,---\, Often unavailable}{Would usually need outside records. \emph{``Result of last year's colonoscopy?''}}
  \rlevel{1 \,---\, Rarely answerable}{Rarely in the source. \emph{``Exact EF on the outside echo?''}}
  \rlevel{0 \,---\, Unanswerable}{Unanswerable from any plausibly available source.}
\end{description}
\end{rubricbox}

\begin{rubricbox}{Redundancy \,---\, is the answer already captured}
\footnotesize
Compares the question against the \emph{current} artifact and prior turns (not the gold reference). High redundancy = the question merely re-confirms known information.
\begin{description}[leftmargin=1.2em,itemsep=1pt]
  \rlevel{5 \,---\, Already documented}{Almost certainly already captured. \emph{``Was a follow-up scheduled?''} when the draft already lists it.}
  \rlevel{4 \,---\, Probably present}{Probably already in the artifact or earlier conversation.}
  \rlevel{3 \,---\, Possibly present}{Unclear whether already captured.}
  \rlevel{2 \,---\, Probably new}{Probably not yet captured.}
  \rlevel{1 \,---\, Genuine gap}{Almost certainly a genuine gap.}
  \rlevel{0 \,---\, Definitely missing}{Brand-new information not present anywhere yet.}
\end{description}
\end{rubricbox}

\paragraph{Tie-breaking.} Across all five rubrics, when in doubt between two adjacent levels the judge is instructed to pick the higher level only if its defining condition (added consequence, more options, an extra sub-answer, clearer presence in the source/artifact) is clearly met; this keeps the distributions from drifting toward 5.

\newpage
\section{Sample Q\&A Traces}
\label{sec:appendix-traces}
We include one full budget-5 trace per task, all drawn from 20\%-degraded variants where DAU gains the most. Each trace shows the baseline ($t{=}0$) draft, the five clarification turns with metric, and the final output.

\subsection{Notes: D2N071-virtassist (ACI-Bench)}
Completeness 47 ($t{=}0$) $\rightarrow$ 69 ($t{=}5$). The patient-facing perturbation reworded the doctor's references to drugs, surgical history, imaging, and exam findings into lay language, leaving the draft note short on specifics. Each clarification recovers a different missing fact, and the priorities are non-decreasing (1$\to$2): the first turn asks the most important gap (medication names), and later turns target progressively more specific refinements.

\subsection{ICD-10: D2N078-aci (ACI-Bench)}
ICD-10 F1 50 ($t{=}0$) $\rightarrow$ 100 ($t{=}3$). Gold codes are \texttt{[S83.411A, I10]}. The clarifications walk a diagnostic decision tree (encounter type $\rightarrow$ anatomy $\rightarrow$ specific structure) typical of ICD-10 specificity; the answer at $t{=}3$ flips the injury code to the correct one. Final prediction: \texttt{[S83.411A, I10]}.

\subsection{Orders: acibench\_D2N177 (MEDIQA-OE)}
Order count 2 ($t{=}0$) $\rightarrow$ 5 ($t{=}5$). Baseline draft captures echocardiogram and lisinopril; subsequent turns recover an A1c lab order, a follow-up visit, and a metformin dose change.

\begin{table*}[ht]
\small
\centering
\begin{tabularx}{\linewidth}{@{}c l c L L r@{}}
\toprule
$t$ & Category & Pri.\ & Question & Answer & Completeness\\
\midrule
1 & medications    & 1 & What specific names of the depression and blood pressure medications is the patient taking? & \textbf{Prozac 20mg and Norvasc} & \textbf{61.1} \\
2 & history\_pmh   & 2 & What type of prior surgery did the patient have? & \textbf{Prior lobectomy} & \textbf{63.9} \\
3 & medications    & 2 & What is the dosage of Norvasc the patient is taking? & No information on this & 61.1 \\
4 & labs\_imaging  & 2 & What specific cardiac imaging study and diagnostic tests were ordered today? & \textbf{Echocardiogram and routine blood work} & 61.1 \\
5 & exam\_findings & 2 & What was the slight heart sound noted on cardiac auscultation? & \textbf{2/6 systolic ejection murmur} & \textbf{69.4} \\
\bottomrule
\end{tabularx}
\caption{Notes Q\&A trace (D2N071-virtassist, ACI-Bench, patient-facing 20\%). Self-reported priorities are non-decreasing (1$\to$2): turn 1 targets the highest-priority gap (specific medication names), and later turns target lower-priority refinements (surgical type, dosage, imaging specifics, exam-finding detail). Contrast with the Orders trace (Table \ref{tab:trace-orders}), where priorities drift higher faster as the list saturates.}
\label{tab:trace-notes}
\end{table*}

\begin{table*}[ht]
\small
\centering
\begin{tabularx}{\linewidth}{@{}c l c L L l@{}}
\toprule
$t$ & Category & Pri.\ & Question / Output & Answer & F1 \\
\midrule
0 & ---              & --- & Pred: \texttt{[M79.661, I10]} & --- & 50.0 \\
1 & encounter\_episode & 1 & Is this the initial encounter for the right knee injury? & Yes, this is initial encounter. & 50.0 \\
2 & type\_subtype      & 1 & Is the injury specifically to the knee joint or the lower leg below the knee? & Knee joint specifically. & 50.0 \\
3 & type\_subtype      & 1 & What specific knee structure is injured: ligament, meniscus, muscle/tendon, or unspecified sprain? & \textbf{Medial collateral ligament sprain} & \textbf{100.0} \\
4 & etiology\_cause    & 1 & Was the fall caused by slipping on ice, water, or another specific surface? & Not documented & 100.0 \\
5 & etiology\_cause    & 1 & Is an external cause code needed: was the slip/fall activity-related or place-of-occurrence documented? & Property work, slip and fall & 100.0 \\
\bottomrule
\end{tabularx}
\caption{ICD-10 Q\&A trace (D2N078-aci, ACI-Bench, dropped-content 20\%). Gold codes \texttt{[S83.411A, I10]}; the structural answer at $t{=}3$ flips the injury code and reaches F1 = 100.}
\label{tab:trace-icd10}
\end{table*}

\begin{table*}[ht]
\small
\centering
\begin{tabularx}{\linewidth}{@{}c l c L L c@{}}
\toprule
$t$ & Category & Pri.\ & Question & Answer & \#orders \\
\midrule
0 & ---                & --- & \textit{Baseline:} echocardiogram; continue lisinopril 20\,mg daily & --- & 2 \\
1 & lab\_order           & 1 & Any lab orders for the elevated hemoglobin A1c? & Hemoglobin A1c in couple months & 3 \\
2 & follow\_up           & 1 & When should Bruce follow up with you? & In a couple months & 4 \\
3 & medication\_details  & 2 & Any changes to metformin dose for diabetes? & \textbf{Increased to 1000mg twice daily} & \textbf{5} \\
4 & other                & 2 & Any referral for therapy for depression? & No therapy referral ordered & 5 \\
5 & other                & 3 & Any dietary recommendations for the leg swelling? & No dietary orders were placed & 5 \\
\bottomrule
\end{tabularx}
\caption{Orders Q\&A trace (acibench\_D2N177, MEDIQA-OE, dropped-content 20\%). Priority drifts upward (1$\to$3) as the list saturates; $t{=}4,5$ are confirmatory negatives.}
\label{tab:trace-orders}
\end{table*}

\noindent Final order list: \emph{Echocardiogram; Lisinopril 20\,mg daily; Metformin 1000\,mg twice daily; Hemoglobin A1c in a couple months; Follow-up visit in a couple months.} Q4 and Q5 are confirmatory negatives, but they do not perturb the existing list, consistent with the small typical $|\Delta_{\text{metric}}|$ on already-saturated artifacts.

\end{document}